\documentclass[sigconf]{acmart}

\AtBeginDocument{%
  \providecommand\BibTeX{{%
    \normalfont B\kern-0.5em{\scshape i\kern-0.25em b}\kern-0.8em\TeX}}}

\usepackage{multirow}
\usepackage{enumitem}
\usepackage{color}
\usepackage{bm}
\usepackage{arydshln}
\usepackage{color,soul}
\usepackage{CJKutf8}
\newcommand\BibTeX{B{\sc ib}\TeX}

\copyrightyear{2020} 
\acmYear{2020} 
\setcopyright{acmcopyright}\acmConference[SIGIR '20]{Proceedings of the 43rd International ACM SIGIR Conference on Research and Development in Information Retrieval}{July 25--30, 2020}{Virtual Event, China}
\acmBooktitle{Proceedings of the 43rd International ACM SIGIR Conference on Research and Development in Information Retrieval (SIGIR '20), July 25--30, 2020, Virtual Event, China}
\acmPrice{15.00}
\acmDOI{10.1145/3397271.3401250}
\acmISBN{978-1-4503-8016-4/20/07}
\begin{document}
\fancyhead{}

\title{GoChat: Goal-oriented Chatbots \\with Hierarchical Reinforcement Learning}


\author{Jianfeng Liu}
\affiliation{%
  \institution{Institute of Computing Technology, Chinese Academy of Sciences. UCAS.}
}
\email{liujianfeng18s@ict.ac.cn}
\author{Feiyang Pan}
\affiliation{%
  \institution{Institute of Computing Technology, Chinese Academy of Sciences. UCAS.}
}
\authornote{Corresponding author.}

\email{panfeiyang@ict.ac.cn}
\author{Ling Luo}
\affiliation{%
  \institution{Institute of Computing Technology, Chinese Academy of Sciences. UCAS.}
}
\email{luoling18s@ict.ac.cn}

\renewcommand{\shortauthors}{Liu, et al.}

\begin{abstract}
A chatbot that converses like a human should be goal-oriented (i.e., be purposeful in conversation), which is beyond language generation. However, existing dialogue systems often heavily rely on cumbersome hand-crafted rules or costly labelled datasets to reach the goals. In this paper, we propose Goal-oriented Chatbots~(GoChat), a framework for end-to-end training chatbots to maximize the long-term return from offline multi-turn dialogue datasets.
Our framework utilizes hierarchical reinforcement learning~(HRL), where the high-level policy guides the conversation towards the final goal by determining some sub-goals, and the low-level policy fulfills the sub-goals by generating the corresponding utterance for response.
In our experiments on a real-world dialogue dataset for anti-fraud in financial, our approach outperforms previous methods on both the quality of response generation as well as the success rate of accomplishing the goal. 
\end{abstract}





\maketitle
\section{Introduction}
Chatbots or human-like dialogue systems play an important role in connecting machine-learning-powered platforms with users.
A chatbot that converses like a human should ideally (1) be capable of fluent language generation, and (2) be driven by long-term goals in a multi-turn conversation. In this paper, we develop a practical unified framework for \textbf{G}oal-\textbf{o}riented \textbf{Chat}bots (GoChat in short), which directly addresses both challenges.

Previous work on chatbots can be roughly categorized into two types, i.e., open-domain chatbots and task-oriented dialogue systems. Open-domain chatbots are not task-specific, therefore can be end-to-end trained on large corpus. With the development of deep sequence-to-sequence~(Seq2Seq) learning, they are modeled as a source to target sequence prediction task. However, they cannot support specific tasks by pursuing the long-term goals. On the other hand, task-oriented dialogue systems~\cite{liu2017iterative,yan2017building} are capable of dealing with domain-specific tasks, e.g., recommending products~\cite{yan2017building} or booking restaurant~\cite{liu2017iterative}.

Recently, some dialogue systems have been proposed to bargain on goods~\cite{lewis2017deal,he2018decoupling}, or recommend products~\cite{kang2019recommendation}.

However, designing such task-oriented dialogue systems often requires lots of hand-crafting, for example, by decomposing the task into sub-tasks with manually-designed rules, or using heavily labelled datasets to build a finite-state machine. Such ad-hoc methods are difficult to scale to complex real-world problems.

In light of these observations, in this paper, we propose the GoChat framework for end-to-end training the chatbot to maximize its long-term return without specific design of the dialogue problem.
We leverage hierarchical reinforcement learning~(HRL) that simultaneously learns a high-level policy to guide the conversation and a low-level policy to generate the response according to the guidance from the high-level policy.

In particular, at each turn of the multi-turn conversation, the high-level policy~(the \emph{manager}) observes the previous conversation as its state, determines a sub-goal as its action, and waits for a reward from the environment representing whether the final goal is accomplished or not. The low-level policy~(the \emph{worker}), on the other hand, observes the state together with the sub-goal from its manager, generates the corresponding utterance for response as its action to fulfill the sub-goal, and receives a reward from the manager.
In this way, the goal is decomposed as a hierarchy of goals, so the chatbot can be trained efficiently in an end-to-end manner.

Our dialogue agent is trained with Advantage Actor-Critic (A2C) \cite{mnih2016asynchronous}, an algorithm widely used in modern reinforcement learning and applications \cite{pan2019pgcr}. We conducted experiments on a goal-oriented dialogue corpus from real-world financial data, where each dialogue has a binary label for whether the goal is achieved or not.
Quantitative indicators as well as human evaluations demonstrate that GoChat outperforms previous methods significantly on both conversation quality and success rate of achieving the goals.

\section{GoChat: Goal-oriented Chatbots}

\begin{figure*}[t]
    \centering
    \includegraphics[width=0.88\textwidth]{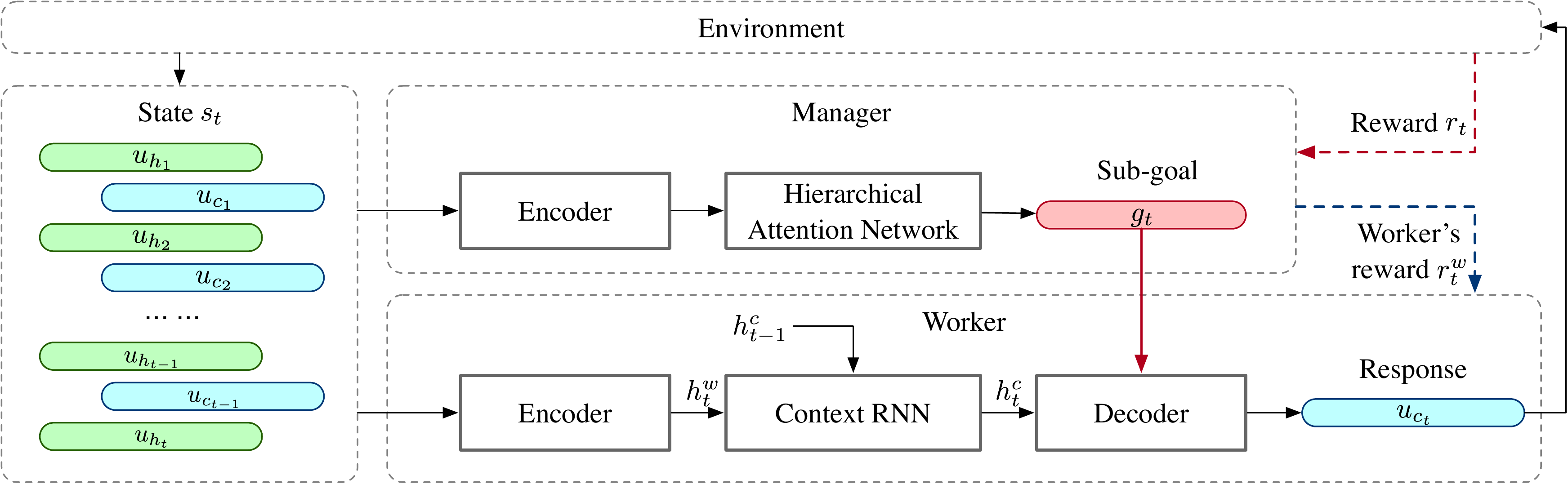}
    \caption{The overall GoChat framework for goal-oriented chatbots with hierarchical reinforcement learning. The manager decides the sub-goals for the worker, and the worker generates the response according to the observations and the sub-goal.}
\label{fig:model}
\end{figure*}

\subsection{Problem formulation}\label{sec:setting}

Recall that our chatbot is intended to accomplish some goal via multi-turn conversations.
Here we denote a complete dialogue with $T$ turns as $\{u_{h_1},u_{c_1},\dots,u_{h_T},u_{c_T}\}$, where $u_{h_t}$ and $u_{c_t}$, $t=1,\dots,T$ stand for the utterances generated by human and chatbot at the $t$-th turn, respectively. Each utterance is a sequence of $n$ tokens~(for sentences with less than $n$ words, we use zero-paddings to complete the sequence).
At the $t$-th turn, given the dialogue of the previous $t-1$ turns as well as the $t$-th utterance produced by human, the chatbot needs to generate $u_{c_t}$ as its response. 
The target of the chatbot is to accomplish the goal through the whole conversation.

We formulate our problem as a Markov Decision Process~(MDP).
At time step $t$, the state is defined as $s_{t}=\{u_{h_1},u_{c_1},\dots,u_{h_t}\}$, and the action is the chatbot's response sequence $u_{c_t}$. For simplicity, we suppose that the reward of accomplishing the goal is given only at the end of the conversation, i.e., $r_T = 1$ if the goal is accomplished, and $r_T = -1$ otherwise. The intermediate rewards are always $0$, i.e., $r_t = 0$ for all $t<T$.

As the outcome can only be observed after a multi-turn conversation, the decision-making process is challenging. 
We adopt Hierarchical Reinforcement Learning~(HRL) to deal with such a long horizon problem with sparse and delayed rewards, and propose the GoChat framework to simultaneously learn the high-level and low-level policy with end-to-end training.

\subsection{The hierarchy of chatbot policies}
The overall framework is illustrated in Figure~\ref{fig:model}.

At the $t$-th turn, given a state $s_{t}=\{u_{h_1},u_{c_1},\dots,u_{h_t}\}$, the high-level policy (the manager) first determines a sub-goal $g_t\in \mathbb{R}^{K}$, which is a one-hot vector drawn from a K-dimensional categorical distribution $g_t \sim \pi_{\psi}(g_t|s_t)$.
Observing the state and the sub-goal, the low-level \emph{worker} outputs the response $u_{c_t}\sim \pi_{\theta}(u_{c_t}| s_t, g_t)$ by an auto-regressive Seq2Seq model, i.e.,
\begin{equation}
\pi_{\theta}(u_{c_t}=(\hat{y}_1,\dots,\hat{y}_n)|s_t, g_t) = \prod_{\tau=1}^n \pi_{\theta}(\hat{y}_\tau|s_t, g_t,\hat{y}_{<\tau})
\end{equation}
Then, the manager receives a reward $r_t$ from the dialogue environment~(described in section~\ref{sec:setting}), while the worker receives a reward $r_t^w$ from the manager~(detailed in section~\ref{sec:worker}).
Next, we introduce the model structures of the manager and the worker.

\subsubsection{Manager}
We adopt the hierarchical attention network (HAN) \cite{yang-etal-2016-hierarchical,luo2018beyond} as the network structure of our manager, which learns the contextual information by sequentially encoding all the previous dialogue turns in the state.
In more detail, each utterance $u_{h_1}, u_{c_1}, \dots, u_{h_t}$ is firstly encoded by a word-level encoder to a hidden vector $h_{h_1}$, $h_{c_1}$, $\dots, h_{h_t}$ with attention mechanism, as the latent representations of utterances.

Next, all the derived sentence vectors $h_{h_1}, h_{c_1}, \dots, h_{h_t}$ are encoded at the dialogue level to form a unified dialogue-level hidden representation $H_t$ also with attention mechanism.
So $H_t$ is a high-level feature to summarize the current conversation, which is used to choose the next sub-goal,
\begin{equation}
g_t \sim \pi_{\psi}(g_t | s_t) = \textrm{Categorical}\big( \mbox{softmax}(W_g^{\top}H_t + b_g) \big),  
\end{equation}
where $W_g$ and $b_g$ are trainable parameters.

\subsubsection{Worker}\label{sec:worker}
The worker generates the response $u_{c_t}$ under the high-level guidance from the manager.
We adopt Variational Hierarchical Recurrent Encoder Decoder (VHRED)~\cite{serban2017hierarchical} as the model structure of the worker, which is a commonly used architecture of multi-turn dialogue systems.
As shown in Figure~\ref{fig:model}, the worker consists of three components for generation, namely the encoder, the context RNN, and the decoder.

At time step $t$, the encoder firstly encodes the human utterance $u_{h_t}$ into an utterance representation $h^w_t$, which is then fed to an higher-level encoder, context RNN, to model the long-term aspects of the conversation. Then, context RNN can produce a dialogue representation $h^c_t$ by taking the $t$-th utterance representation $h^w_t$ and the dialogue representation $h^c_{t-1}$ from last conversation as input.
Next, $h^c_t$ is fed to a fully connected layer to predict a mean $\bm{\mu}$ and a variance $\bm{\sigma}$ of a diagonal Gaussian distribution as the variational posterior approximation for a latent variable $z_t$.
Then, the latent $z_t$ is sampled from the posterior using the reparameterization trick, i.e., $z_t = \bm{\mu} + \bm{\sigma} \circ \bm{\epsilon}$, where $\bm{\epsilon}\sim \mathcal{N}(\bm{0}, \bm{1})$. Therefore, the latent can be seen as drawn conditioned on the state: $\bm{z}_t \sim \mathcal{N}(\bm{\mu}(s_t),\bm{\sigma}(s_t))$.

Finally, the sub-goal from the manager influences the generation process by perturbing the latent $\bm{z}_t$.
We get an augmented latent variable by simply concatenating the latent $z_t$ with $g_t$, i.e., $\tilde{z}_t = [z_t, g_t]$.
Finally, the output utterance ${u}_{c_t}=[\hat{y}_1,\dots,\hat{y}_n]$ can be generated by the sequential decoder with the input of $\tilde{z}_t$ and $h^c_t$.

It is challenging to design a reasonable reward function to measures how well the worker has achieved its sub-goal, especially when our agent is trained on off-policy data.
In our framework, we use the following way to directly construct such reward with the offline dataset. First, before the end-to-end training, each (source, target) pair $(s, u_{c})$ is assigned with an \emph{offline} sub-goal vector using unsupervised clustering, so that the dataset becomes a set of tuples in the form of $(s, g, u_{c})$. Then, during training the RL agent, given the state $s_t$ and the sub-goal $g_t$ from the manager, we find the $k$ nearest neighbors of $(s_t, g_t)$ from the offline dataset and get $k$ reference responses. Finally, we compute the average BLEU score between our generated $u_{c_t}$ and the $k$ references as the reward $r_t^w$.

\subsection{End-to-end training the chatbot}\label{sec:training}
In this part, we describe the training methods of our frameworks. 

To accelerate learning, we first use a pre-train step to warm start the agent. Recall that we have generated the offline sub-goals for each source-target pair, we can pretrain the manager and the worker with supervised Seq2Seq learning.

Then, we use a simulation-based environment to end-to-end optimize the goal-oriented chatbot. Following previous work on model-based RL \cite{pan2019pome} and RL-for-dialog~\cite{saleh2019hierarchical}, we simulate the interactive simulator with a user model. Each conversation is initialized with some starting utterances randomly sampled from the training set. Then, our agent chats with a learned user model. The user model has the same network structure with our agent, and is trained with supervised Seq2Seq learning on the dataset to predict the human response and is fixed after training. We limit the multi-turn simulation dialogues to $m$ turns~($2m$ utterances), and $m$ is up to the practical application scenarios, to prevent the agents from going off-topic on overlong conversations.

During training the HRL agent, we simultaneously optimize the manager and worker to maximize the following objective function
\begin{equation}\label{eq:actor_loss}
J(\pi_\psi, \pi_\theta) = \mathbb{E}_{g_t\sim \pi_\psi(\cdot|s_t), u_{c_t}\sim \pi_\theta(\cdot|s_t, g_t)}\big[\sum_{t=1}^{T} \alpha r_{t} + \beta  r^w_{t} \big]
\end{equation}
where the expectation is over conversations induced by the policies, and $\alpha, \beta$ are hyper-parameters to balance the two levels of rewards.

To maximize $J(\pi_\psi, \pi_\theta)$, we use the Advantage Actor-Critic~(A2C) \cite{mnih2016asynchronous} algorithm, where the policies are updated according to the estimated advantage function of the chosen actions. E.g., for the manager, $\hat A(s_t, g_t) = R_t(s_t, g_t) - \hat V_\phi(s_t)$, where $R_t=\sum_{\tau=t}^T r_t$ is the long-term return and $\hat V_\phi(s_t)$ is the estimated value function of state $s_t$ with parameters $\phi$. 
Then, the surrogate loss for policy learning is
\begin{equation}
    L_{\pi} = -\hat{\mathbb{E}}_t\big[\sum_{t=1}^{T} \alpha \hat A_{t} \log \pi_\psi(g_t|s_t) + \beta  A^w_{t}\log \pi_\theta(u_{c_t} | s_t, g_t)\big]
\end{equation}
where $\hat{\mathbb{E}}_t[\cdot]$ denotes the sample mean over rollout dialogues induced by the policies. 

We adopt the hierarchical attention network~(HAN) as the value network $\hat V_\phi(s_t)$, which is trained to minimize the following squared error according to the Bellman equation
\begin{equation}\label{eq:critic_loss}
L_{v} = \frac{1}{2}\big(r_{t+1} + \gamma V_{\phi}^{\textrm{targ}}(s_{t+1}) - V_{\phi}(s_{t})\big)^2
\end{equation}
where $V_{\phi}^{\textrm{targ}}$ is the target value network of A2C and $\gamma=0.99$ is the discount factor.
Finally, we have a unified loss 
$\mathcal{L} = L_{\pi} + L_{v}.$

\section{Experiments}
\subsection{Experimental set-up} \label{data}
We conduct experiments on a anti-fraud dialogue system on a large-scale FinTech platform. Our task is to train a chatbot for assisting the security officers to have a conversation with suspected scammers.
The goal of the chatbot is to find evidences of the scammers' fraud, e.g., to get to know their account ID so that the security officers can take further investigation.
We collect a dataset of over four thousand historical dialogues between the suspected scammers and the security officers, where each dialogue has about 20 turns in average and is only labeled with a binary outcome, indicating whether the goal is accomplished.
The data is anonymized without any physical person's privacy information.

Given such dataset, we compare GoChat against state-of-the-art multi-turn dialog methods, including HRED~\cite{serban2016building}, VHRED~\cite{serban2017hierarchical}, WSeq~\cite{tian2017make}, HRAN~\cite{xing2018hierarchical}, and ReCoSa~\cite{zhang2019recosa}. We do not compare other task-oriented approaches which require detailed annotations on each query-response pairs or dialogue states.

\begin{table}[]
\caption{Quantitative results. D-1,-2 stands for Distinct-1,-2.}
\label{tab:result}\vspace{-1mm}
\begin{tabular}{lcccccc}
\toprule
\multirow{2}{*}{Method} & \multicolumn{3}{c}{v.s. GoChat} & \multirow{2}{*}{BLEU} & \multirow{2}{*}{D-1} & \multirow{2}{*}{D-2} \\ \cline{2-4}
        & Win(\%)   & Tie(\%)  & Lose(\%)   &                        \\ \hline
HRED    &    17.3   &    30.2  &  52.5 &     9.21      &  .051      &  0.407  \\
VHRED   &    22.6   &    40.0  &  37.4 &  9.32      &  .053 &   0.419          \\
WSeq    &    9.5    &    15.3  &  75.2 &      6.43  & .047 &   0.372    \\
HRAN    &    14.9   &    12.8  &  72.3 &    7.18   &  .048 & 0.384    \\
ReCoSa  &    22.1   &    32.6  &  45.3 & 9.01   &  .050 & 0.415 \\ 
\midrule
GoChat & - & - & - & \textbf{9.70} &  \textbf{.061} &  \textbf{0.479}   \\ 
\bottomrule
\end{tabular}
\end{table}\label{tab:case}

\begin{table*}[] \small
\caption{Case study: comparison of generated dialogue from our GoChat and the most competitive baseline, VHRED.}\vspace{-2mm}
\label{tab:case}
\begin{tabular}{cll}
\toprule
& \textbf{Agent} & \textbf{Utterance} \\ \hline

\multirow{6}{*}{\textbf{Case~(A)}} & \multirow{2}{*}{Human (Scammer):} & \begin{CJK*}{UTF8}{gkai} 亲，您好！欢迎您来到企业招聘兼职中心 ，请问您是否需要兼职？\end{CJK*} \\ 
& & (Welcome to the Enterprise Part-Time Recruitment Center. Do you need a part-time job?) \\
& \multirow{2}{*}{VHRED (Officer):} &\begin{CJK*}{UTF8}{gkai} 是的 \end{CJK*} \\ 
& & (Yes. )\\
& \multirow{2}{*}{GoChat (Officer):} &\begin{CJK*}{UTF8}{gkai} 需要。有什么兼职可以做呢？ \end{CJK*} \\ 
& & (Yes, I need. What kinds of part-time jobs do you have? )\\
\hline

\multirow{6}{*}{\textbf{Case (B)}} & \multirow{2}{*}{Human (Scammer):} & \begin{CJK*}{UTF8}{gkai} 你是谁推荐来贷款的?\end{CJK*} \\ 
& & (Who recommended you here for the loan?) \\
& \multirow{2}{*}{VHRED (Officer):} &\begin{CJK*}{UTF8}{gkai} 从一个朋友那里 \end{CJK*} \\ 
& & (From one of my friends.)\\
& \multirow{2}{*}{GoChat (Officer):} &\begin{CJK*}{UTF8}{gkai} 就是打游戏的时候，人家说可以办理贷款的。请问可以办理嘛？ \end{CJK*} \\ 
& & (When I was playing a game, someone said I could apply for a loan. Can I?)\\
\bottomrule
\end{tabular}
\end{table*}

\subsection{Implementation details and evaluation}

In our experiments, we limit the maximum length of dialog to $m=20$ turns~($40$ utterances) and adopt $500$-size word embeddings for all methods.
The hidden size of word-level and dialog-level encoders in HAN~(both the manager and the critic) are $500$ and $50$, respectively.
The configuration of the worker follows VHRED \cite{serban2017hierarchical}.

We firstly pretrain the worker on $(s_t,g_t,u_{c_t})$ tuples detailed in Section~\ref{sec:worker} with supervised Seq2Seq learning, where the offline sub-goal $g_t$ is given by Latent Dirichlet allocation~(LDA) into $14$ categories~(sub-goals).
We set the hyper-parameters in Eq.~\ref{eq:actor_loss} as $\alpha=1,\beta=0.001$.
The agent is optimized with Adam \cite{kingma2014adam}, and trained for $10$ epochs with a learning rate of $0.001$.

We adopt human evaluation to examine the utterances for achieving goals.
In detail, eight annotators are asked to compare the simulated response between our GoChat and each baseline given the same context.
Annotators would give it a tie if both our method and baselines fail/succeed to generate utterances which are beneficial to achieve the goal, otherwise they would annotate a win/lose.
We also use BLEU and distinct~\cite{li-etal-2016-diversity} to evaluate the quality and the diversity of the generated responses, respectively.

\subsection{Quantitative results} \label{sec:res}
We compare our model with five baselines for multi-turn dialogue, and report the quantitative indicators and human evaluation in Table~\ref{tab:result}.
From the perspective of generation indicator, our proposed GoChat get the highest score on BLEU~($9.70$), distinct-1~($.061$) and distinct-2~($0.479$), which demonstrates our GoChat can generate more high-quality and varied utterances than baseline methods.
From the human evaluation, we observe that GoChat outperforms all the baselines on generating utterances for achieving the goal. Specifically, GoChat is preferred at rates of $35.2\%$, $14.8\%$, $65.7\%$, $57.4\%$, $23.2\%$ against HRED, VHRED, WSeq, HRAN and ReCoSa, respectively.
We conjecture it might be the simultaneous training of high-level and low-level policy that our GoChat better models long-term rewards the goals and produces high-quality responses.
\subsection{Qualitative results}

According to Table~\ref{tab:result}, VHRED is the most competitive method. To compare, we present two representative cases in Table~\ref{tab:case}. Observing these cases, both two methods can generate appropriate responses to the inquiries of the scammer given same dialogue contexts.
However, our model is more likely to raise questions and generate more diverse sentences. Specifically, in both cases, the utterances from our GoChat, i.e, ``What kinds of part-time jobs do you have?'', ``Can I?'', are more proactive.
Such responses are more likely towards the ultimate goal in a long run.
While case~(B) demonstrates that our GoChat can generate more diverse and human-like responses, i.e., ``When playing games'', which is very convincing and might help to get rid of the scammers' doubts in anti-fraud dialogue tasks.
\section{Conclusion}
In this paper, we propose an end-to-end framework GoChat for goal-oriented dialogue systems based on hierarchical reinforcement learning without cumbersome hand-crafting.
It uses a high-level policy to determine the sub-goal to guide the conversation towards the final goal and a low-level worker to generate corresponding responses. Experiments conducted on a real-world dialogue dataset verified the effectiveness of our GoChat.
\section*{acknowledgement}
The research work is supported by the National Natural Science

\noindent Foundation of China under Grant No. 61976204, U1811461, the Project of Youth Innovation Promotion Association CAS. This work is also funded in part by Ant Financial through the Ant Financial Science Funds for Security Research.
\bibliographystyle{ACM-Reference-Format}
\bibliography{sample-base}
\end{document}